\documentclass[conference]{IEEEtran}
\usepackage{cite}
\usepackage{amsmath,amssymb,amsfonts}
\usepackage{algorithmic}
\usepackage{graphicx}
\usepackage{textcomp}
\usepackage{xcolor}
\usepackage[caption=false]{subfig}

\usepackage{url}
\usepackage{multicol}
\usepackage{multirow}
\usepackage{color}
\usepackage{hhline}
\usepackage{pifont}
\usepackage{comment}
\usepackage[ruled,vlined]{algorithm2e}
\definecolor{mynicegreen}{RGB}{11,102,35}

 \newcommand{\squishlist}{
	\begin{list}{$\bullet$}
		{ \setlength{\itemsep}{0pt}
			\setlength{\parsep}{3pt}
			\setlength{\topsep}{3pt}
			\setlength{\partopsep}{0pt}
			\setlength{\leftmargin}{1.5em}
			\setlength{\labelwidth}{1em}
			\setlength{\labelsep}{0.5em} } }
	
	\newcommand{\squishlisttwo}{
		\begin{list}{$\bullet$}
			{ \setlength{\itemsep}{0pt}
				\setlength{\parsep}{0pt}
				\setlength{\topsep}{0pt}
				\setlength{\partopsep}{0pt}
				\setlength{\leftmargin}{2em}
				\setlength{\labelwidth}{1.5em}
				\setlength{\labelsep}{0.5em} } }
		
		\newcommand{\squishend}{
	\end{list}}
	
\def\BibTeX{{\rm B\kern-.05em{\sc i\kern-.025em b}\kern-.08em
    T\kern-.1667em\lower.7ex\hbox{E}\kern-.125emX}}

\begin{document}

\title{An Effective, Robust and Fairness-aware\\Hate Speech Detection Framework
}

\IEEEoverridecommandlockouts
\IEEEpubid{\makebox[\columnwidth]{978-1-6654-3902-2/21/\$31.00~\copyright2021 IEEE \hfill} \hspace{\columnsep}\makebox[\columnwidth]{ }}

\author{\IEEEauthorblockN{Guanyi Mou}
\IEEEauthorblockA{\textit{Worcester Polytechnic Institute} \\
100 Institute Rd, Worcester, MA, 01605 \\
gmou@wpi.edu}
\and
\IEEEauthorblockN{Kyumin Lee}
\IEEEauthorblockA{\textit{Worcester Polytechnic Institute} \\
100 Institute Rd, Worcester, MA, 01605 \\
kmlee@wpi.edu}
}

\maketitle
\IEEEpubidadjcol

\begin{abstract}
With the widespread online social networks, hate speeches are spreading faster and causing more damage than ever before. 
Existing hate speech detection methods have limitations in several aspects, such as handling data insufficiency, estimating model uncertainty, improving robustness against malicious attacks, and handling unintended bias (i.e., fairness).
There is an urgent need for accurate, robust, and fair hate speech classification in online social networks. To bridge the gap, we design a data-augmented, fairness addressed, and uncertainty estimated novel framework.
As parts of the framework, we propose Bidirectional Quaternion-Quasi-LSTM layers to balance effectiveness and efficiency. To build a generalized model, we combine five datasets collected from three platforms. Experiment results show that our model outperforms eight state-of-the-art methods under both no attack scenario and various attack scenarios, indicating the effectiveness and robustness of our model.
We share our code along with combined dataset for better future research\footnote{\url{https://github.com/GMouYes/BiQQLSTM_HS}}.
\end{abstract}


\section{Introduction}
\label{Sec:Introduction}

Hate speech has long been causing annoying disturbances and damage to many people's lives by misleading the topic trends, shaping bias and discrimination, aggregating and aggravating conflicts among different religious/gender/racial groups. With the rapid growth of online social networks, hate speech is spreading faster and affecting a larger population than ever before in human history\cite{laub2019hate}.
Therefore, quickly and accurately identifying hate speech becomes crucial for mitigating the possible conflicts, keeping a harmonic and healthy online social environment, and protecting our society's diversity.

Researchers have proposed various methods for detecting hate speech~\cite{nobata2016abusive,djuric2015hate,badjatiya2019stereotypical,liu2019fuzzy,chowdhury2019arhnet,zhang2018detecting}. However, existing approaches in hate speech detection have the following limitations.

First, the prior work mostly used insufficient data, and the quality of data was varied. For example, used/shared hate speech datasets contain a limited amount of data~\cite{degibertetal2018hate,wulczyn2017ex,alrehili2019automatic}.
The definition of ``hate speech'' varies across works, and they inevitably affect the researchers' different methodologies and criteria for collecting, filtering, and labeling data~\cite{ousidhoum2019multilingual,davidson2017automated,waseem2016hateful,elsherief2018hate}.
These may cause unintended bias and errors inside the datasets. In addition, some researchers only focused on data obtained from a single platform/website, which puts constraints on the model's generalizability to other platforms.
Second, the importance of balancing between effectiveness and efficiency in model designs is often ignored. Prior works often only focused on improving effectiveness.
Third, prior works in hate speech detection did not thoroughly test and adapt data augmentation techniques (e.g., character-level perturbation, word-level synonym replacement, natural language generation) toward building robust models against various attacks and text manipulation.
Fourth, fairness in hate speech detection models is less addressed, although fairness has become an important issue in other domains such as sentiment classification~\cite{garg2019counterfactual}.
Lastly, the existence of predictive uncertainty of the hate speech detection models shall be taken better care of. We need a mechanism to balance the bias and variance of predictions.

In this paper, we embrace the hate speech definition as ``abusive speech targeting specific group characteristics''~\cite{warner2012detecting} to take both generality and specificity of hate speeches into consideration. We further propose a novel framework for hate speech detection to overcome the aforementioned limitation and challenges. In particular, we combine five datasets collected from three platforms to build a generalized model. The framework consists of our proposed Bidirectional Quaternion-Quasi-LSTM (BiQQLSTM) layers to balance effectiveness and efficiency. To handle the fairness and predictive uncertainty of the model, our loss function narrows the gap between original texts and their counterfactual logit pairs and leverages tunable parameters for estimating true uncertainty. To build a robust model against various text manipulation/attacks, we adapt and customize existing data augmentation techniques for the hate speech domain.

The major contributions of our work are as follows:
\squishlist
\item We propose a BiQQLSTM framework, which customizes the original Bidirectional Quasi-RNN by replacing real-valued matrix operations with quaternion operations. The quaternion operations help improving effectiveness, and the quasi-RNN helps reducing running time (i.e., improving efficiency).
\item We incorporate fairness into our framework to mitigate unintended bias and further estimate the model's predictive uncertainty to further improve performance.
\item We propose an augmentation strategy with: (i) a \textbf{generative method} to resolve \textit{data insufficiency}; (ii) an \textbf{optimal perturbation based augmentation combinations} for better \textit{model robustness under malicious attacks}; and (iii) a \textbf{filtering mechanism} to improve the augmented data quality and reduce \textit{data uncertainty and injected noise}.
\item Extensive experiments show that our proposed framework outperforms 8 state-of-the-art baselines with 5.5\% improvement under no attack scenario; and up to 3.1\% improvement under various attack scenarios compared with the best baseline, confirming the effectiveness and robustness of our approach.
\squishend

\section{Related Work}
\label{Sec:RelatedWork}



\subsection{Hate Speech Detection}
Both \cite{warner2012detecting} and \cite{waseem2016you} conducted in-depth analysis of hate speech. Arango et al.\cite{arango2019hate} analyzed a model validation problem of hate speech detection. For classification tasks, Nobata et al.\cite{nobata2016abusive} tried various types of features and reported informative results. Recent papers leveraged CNNs, LSTMs and attention mechanisms for better detection results~\cite{badjatiya2017deep,gamback2017using,liu2019fuzzy,chowdhury2019arhnet,badjatiya2019stereotypical,zhang2018detecting}. Djuric et al.\cite{djuric2015hate} experimented on using paragraph-level embeddings for hate speech detection. Mou et al.\cite{mou2020swe2} leveraged both LSTMs on word/word-piece embeddings and CNNs on character/phonetic embeddings. Intentional manipulation made by hate speech posters can possibly evade the prior hate speech detection methods~\cite{sun2020adv,garg2020bae,li2020bert,li2020contextualized}.

Researchers\cite{ousidhoum2019multilingual,davidson2017automated, waseem2016hateful, elsherief2018hate, degibertetal2018hate, wulczyn2017ex} released their annotated hate speech datasets in public. We made use of the latter five datasets in our research and described them in detail in Section~\ref{Sec:Dataset}. Others\cite{mathew2019thou, chung2019conan} provided counter speech datasets for better analysis of hate speech. Aside from these public datasets, Tran et al.\cite{tran2020habertor} leveraged datasets from Yahoo News and Yahoo Finance for building a hate speech detector. 
Most researchers only used a single-platform dataset (mainly on Twitter), so their models may not be generalized well for detecting hate speech on other platforms. Unlike the prior work, we trained our model with data from three platforms to make it more generalizable.

\subsection{Data Augmentation}
\label{subsec:relatedaugmentation}
Data augmentations mainly contribute to make a model \emph{generalizable} and \emph{robust} by providing data varieties. Unlike our work, existing methods usually emphasized either side (i.e., generalizable vs. robust), but very few of them addressed both perspectives in one framework.
Generally speaking, improving generalization will consequently improve model performance in a standard/no attack scenario. Improving robustness 
can sometimes even hurt performance. However, it will help mitigate the impact of malicious attacks (manipulations).

Augmentations can vary from perturbation methods to generative methods. Perturbation methods
creates (usually) small and controllable changes to the given original samples\cite{miller1995wordnet,alzantot2018generating,ribeiro2020beyond,jiao2020tinybert,garg2020bae,wu2019conditional,xie2020unsupervised,luque2019atalaya,li2019textbugger,pruthi2019combating,shen2020simple,weizou2019eda,coulombe2018text,rizos2019augment}. 
They usually have low costs and are easy to scale. However, automatic perturbations will inevitably change the original context, which sometimes hurts the contents' quality. Changes in some significant words can even cause label flipping problems~\cite{yi2021reweighting}.
Generative methods enable a deep learning network to capture the pattern of a given corpus of the dataset and then generate similar data from a given starter (or nothing)~\cite{wullach2020towards,radford2019language,cao2020hategan,anaby2020not,kumar2020data}.


To resolve these problems, we incorporated five representative perturbation methods and explored their optimal combinations for improving robustness~\cite{li2019textbugger,miller1995wordnet,wang2015s,ribeiro2020beyond,weizou2019eda}. We also trained a fine-tuned task-specific generative model for improving general performance. Lastly, we designed a filtering mechanism on these augmentation methods to improve data quality, reducing injected noise, and minimize data uncertainty.

\subsection{Fairness and Uncertainty Estimation}
\label{subsec:fairness}
Fairness has been recently addressed and discussed in various domains such as business activities~\cite{frezal2019fairness}, recommendation systems~\cite{wang2020fairness,patro2020incremental}, and general language models~\cite{nissim2020fair}. Early literature treated fairness as ``equality of opportunity'' in general classification problems~\cite{hardt2016equality,zhang2018equality}, and researchers explored theoretical proofs and methodologies for improving fairness~\cite{beutel2017data,zhang2018mitigating,celis2019classification,dixon2018measuring}.
Researchers in other domains proposed various uncertainty estimation approaches such as ensemble models~\cite{lakshminarayanan2017simple}, dropout methods~\cite{gal2016dropout} and probability estimation~\cite{loquercio2020general,borsuk2004bayesian}. However, researchers in the hate speech domain did not pay much attention to fairness and uncertainty estimation.

\begin{figure*}[ht]
\centering
    \includegraphics[width=.75\linewidth]{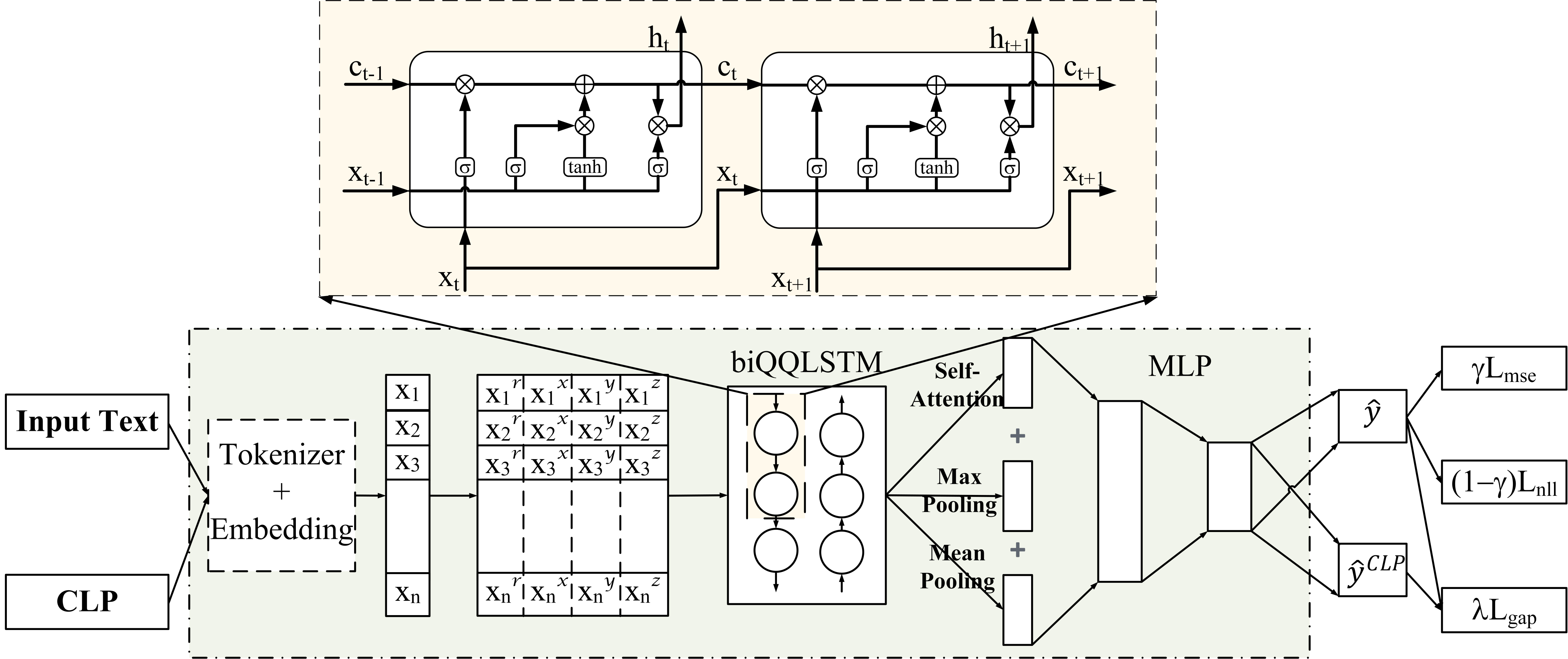}
    \caption{The overview of our framework.}
    \label{fig:network}
    \vspace{-10pt}
\end{figure*}

\section{Preliminary}
\label{Sec:Background}
In this section, we cover background on Quaternion algebra and networks and Quasi-RNNs we used to design our model.

\subsection{Quaternion Algebra and Networks}
\label{subsec:quaternion}
We value Quaternion with its capability of 1) capturing internal dependencies among network features; and 2) reducing learnable free parameters, thus reducing possible network overfitting on the data. We adapt them for the hate speech detection task for better-facilitating network performance. Below, we introduce 1) Quaternion networks; and 2) the notations of quaternions and operations among the inputs, outputs, and weights of our proposed BiQQLSTM.

Quaternion networks have been applied to computer vision~\cite{gaudet2018deep} and human motion classification~\cite{pavllo2018quaternet,pavllo2019modeling}, where rotations to the 3D image or 3D space coordination were common and essential operations. It gained popularity in recommender systems~\cite{ijcai2019-599,tran2020quaternion}, and the natural language processing domain~\cite{tay2019lightweight,parcollet2016quaternion,tran2020habertor}. Parcollet et al.\cite{parcollet2019quaternion} especially proposed quaternion recurrent neural networks for speech recognition.


\smallskip A quaternion $Q$ is a complex number defined in four dimensional space
\begin{equation}
\small
    Q = r\textbf{1}+x\textbf{i}+y\textbf{j}+z\textbf{k}
\end{equation}
where $r,x,y,z$ are real numbers, and $\textbf{1, i, j, k}$ are the quaternion unit basis. Specifically, for the imaginary parts, we have
\begin{equation}
\small
    \textbf{i}^2 = \textbf{j}^2 = \textbf{k}^2 = \textbf{ijk} = -1
\end{equation}
The conjugate $Q^*$ of $Q$ is represented as
\begin{equation}
\small
    Q^* = r\textbf{1}-x\textbf{i}-y\textbf{j}-z\textbf{k}
\end{equation}
The norm of Q is represented as
\begin{equation}
\small
    |Q| = \sqrt{r^2+x^2+y^2+z^2}
\end{equation}
Thus the unit quaternion $Q^\triangleleft$ can be written as
\begin{equation}
\small
    Q^\triangleleft = \frac{Q}{\sqrt{r^2+x^2+y^2+z^2}}
\end{equation}
Given two quaternions $Q_1=r_1\textbf{1}+x_1\textbf{i}+y_1\textbf{j}+z_1\textbf{k}$ and $Q_2=r_2\textbf{1}+x_2\textbf{i}+y_2\textbf{j}+z_2\textbf{k}$,
the Hamilton product of $Q_1$ and $Q_2$ encodes latent dependencies between latent features
\begin{equation}
\small
\begin{aligned}
    Q_1 \otimes Q_2 ={} & (r_1r_2-x_1x_2-y_1y_2-z_1z_2)\textbf{1}+ \\
                      & (r_1r_2+x_1x_2+y_1y_2-z_1z_2)\textbf{i}+ \\
                      & (r_1r_2-x_1x_2+y_1y_2+z_1z_2)\textbf{j}+ \\
                      & (r_1r_2+x_1x_2-y_1y_2+z_1z_2)\textbf{k}
\end{aligned}
\end{equation}
The Hamilton product captures the internal latent relations within the features encoded in a quaternion, so it provides better network performance~\cite{parcollet2019quaternion,tran2020habertor}.

\noindent For an activation function $\alpha$ on real values, the corresponding activation $\beta$ on a quaternion $Q$ can be defined as 
\begin{equation}
\small
    \beta(Q) = \alpha(r)\textbf{1}+\alpha(x)\textbf{i}+\alpha(y)\textbf{j}+\alpha(z)\textbf{k}
\end{equation}
\noindent To transform a real-valued vector $V \in R^{4n}$ into a Quaternion, we simply divide it into 4 parts, $r,x,y,z$ where each part is in $R^n$. To transform a Quaternion representation into a real-valued vector, we concatenate all 4 parts together.

\subsection{Quasi-RNNs}
\label{subsec:quasi}
Quasi-Recurrent Neural Networks~\cite{bradbury2017quasi} (QRNN) were proposed to improve traditional RNN structures' efficiency. However, this approach still keeps most of the merits of the recurrent designs. The authors of QRNN tested their model across multiple tasks such as sentiment classification and neural machine translation and got comparable results to the state-of-the-art LSTM design.

We value both effectiveness and efficiency in our network design. While quaternions effectively capture the internal dependencies and reduce learnable free parameters, they would cause longer training time due to more computation operations. Therefore, we adapt Quasi-RNNs into our framework to speed up general recurrent designs. We are the first to test its effectiveness in the hate speech detection domain and customize it with quaternion operations to the best of our knowledge.

\section{Our Proposed Framework}
\label{Sec:Framework}
Fig.~\ref{fig:network} depicts our proposed framework. We describe each part in the following subsections.
\subsection{Data preprocessing}
We replace sensitive personal information with specific unique tokens to protect user privacy and avoid bias against a person. In particular, each link/URL and each mention (e.g., @bob) are replaced with ``URL'' and ``MENTION'', respectively. Domain-specific tags such as ``rt'', ``fav'' are removed to make a model generalizable. After this preprocessing, we expand necessary contractions related to custom conventions and remove punctuation.

\subsection{Generating CLP during training time}
\label{subsec:clp}
We believe similarly sensitive entities deserve the same level of respect in hate speech detection. Specifically, assume we have a sensitive entity pair A and B deemed equally important. Then, in general, replacing B with a hateful speech targeting A should also be treated as hate speech and vice versa. We thus introduce an explicit fairness module into our network.

We follow \cite{garg2019counterfactual} for generating counterfactual logit pairs (CLP).
The ultimate goal of CLP is to minimize the classification difference between the original text and the CLP through the network.
For example\footnote{\textbf{\label{footn:Disclaimer}We minimized showing hate speech examples. They do not represent the views of the authors.}}:
\squishlist
\item Hate speech from a public dataset: \textit{``The \textbf{lesbian} student will probably find ...''}
\item One possible CLP: \textit{``The \textbf{homosexual} student will probably find ...''}
\squishend
In this example, the target transformed from \textbf{lesbian} to \textbf{homosexual} people. However, one can easily tell that both the original speech and the generated CLP can be viewed as apparent hate speech.

Such a CLP of the input text (i.e., a message) is created in each training iteration by replacing certain sensitive named entities with their equally essential counterfactuals. Both the original input text and the CLP go through the same deep learning architecture, and the architecture will output predictive probabilities for the two strings (i.e., $\hat y$, $\hat{y}^{CLP}$). A special loss described in Section~\ref{Subsec:LossFunction} will try to minimize the difference between these two probabilities in detail.

\vspace{-5pt}
\subsection{Deep Learning Classifier}
\label{sec:modelDesign}
We design a novel deep learning classifier named BiQQLSTM, utilizing both quasi-rnns and quaternion operations. As mentioned in Section~\ref{subsec:quaternion} and Section~\ref{subsec:quasi}, quaternion operations effectively capture internal dependencies among the features and reduce the number of learnable parameters, thus preventing the model from overfitting. Meanwhile, quasi-rnns accelerate training. By combining both components, we aim to improve our model's effectiveness and efficiency. To the best of our knowledge, no prior works combine both quaternion and quasi-rnns. We are also the first to introduce quasi-rnns to the hate speech classification domain.

Fig.~\ref{fig:network} shows the input, output, network components as well as loss calculations. Given that an input text $T$ is represented as a string sequence, our framework will predict whether it is a hate speech or a legitimate speech.
The input text is first tokenized and later goes through an embedding layer. We leveraged word-piece tokenizer and BERT as our embedder.
After tokenization, $T$ is segmented into a $n$ word-pieces sequence, represented as $W$. The matrix of the embedding vectors is denoted by $X_i \in R^{n \times d}$, where $n$ is the sequence length, and $d$ is the dimension of the embedding vector.
\begin{equation}
\small
\begin{aligned}
    & W = (w_1, w_2, ..., w_n) = tokenize(T) \\
    & [X_i^r, X_i^x, X_i^y, X_i^z] = X_i = embed(W)
\end{aligned}
\end{equation}
As mentioned in Section~\ref{subsec:quaternion}, $X_i$ is then transformed into quaternions $[X_i^r, X_i^x, X_i^y, X_i^z]$ and fed to our proposed multi-layer Bidirectional Quaternion-Quasi-LSTM (BiQQLSTM) for capturing and extracting useful information out of the rich embeddings for classification (in practice, we found two layers performed the best).
The BiQQLSTM is made up of the backbone of Quasi-LSTM, but all of its inputs, weights, outputs are replaced as quaternions. All matrix operations, including dot products and activations, are also replaced with quaternions operations in four-dimensional complex space.

More specifically, the Bi-Quaternion-Quasi-LSTM change the gates in classic Bi-LSTMs, from recurrent designs on both inputs and hidden states to convoluted designs on solely inputs, such that the representations become:
\begin{equation}
\small
F, O, I = \sigma([W_f, W_o, W_i] * X)
\end{equation}
where $*$ means convolution operation, $F, O, I$ represents forget gate, output  gate and input gate. Meanwhile, the cell state and the hidden state still preserve their recurrent manner:
\begin{equation}
\small
    \begin{aligned}
    & c_t = f_t \odot c_{t-1} + i_t \odot z_t \\
    & h_t = O_t \odot c_t
    \end{aligned}
\end{equation}
In this way, computations are accelerated in gates as convolutions can be computed in parallel, where local dependencies (interpreted as n-grams) are captured, while the long-term dependencies remain in cell states and hidden states. Note that all the above notations are actual quaternion matrices rather than real-value matrices.

We show the visualization of details of each recurrent cell's design in the upper part of Fig.~\ref{fig:network}.
In the figure, $\sigma$ and $tanh$ are activation functions, $\otimes$ represents element-wise multiplication, and $\oplus$ represents element-wise addition.
Another difference for the Bi-QQLSTM against traditional LSTM is: the $tanh$ activation originally applied on $c_t$ in the calculation of $h_t$ is omitted. We tested and compared the performance between using VS. omitting it and found no significant difference. Thus for model simplicity, we opt in favor of not applying the extra $tanh$ function.

In this way, we combine quaternion operations' performance boost and Quasi-LSTM's faster running time. The BiQQLSTM layers' final output is denoted as $[X_o^r, X_o^x, X_o^y, X_o^z]$, it is then transformed from quaternions back to real values $X_o \in R^{n \times 2*4h}$, where $4h$ is the hidden dimension of BiQQLSTM per direction for all four parts in a quaternion.
\begin{equation}
\resizebox{0.8\hsize}{!}{%
    $X_o = [X_o^r, X_o^x, X_o^y, X_o^z] = BiQQLSTM([X_i^r, X_i^x, X_i^y, X_i^z])$%
    }
\end{equation}
$X_o$ is then fed to three different mechanisms for reducing noise and extracting useful information: mean pooling, max pooling, and self-attention. The self-attention automatically learns the importance of each word and addresses it differently, while the max pooling and the mean pooling keep the statistical information across all words. The results of the three extraction methods are concatenated to form a vector $V \in R^{2*3*4h}$. Similar approaches were effective in early language models and applications~\cite{howard2018universal,mou2020swe2, mou2020malicious}.
\begin{equation}
\small
    \begin{aligned}
    & V_{avg} = AVG\_Pool(X_o) \\
    & V_{max} = MAX\_Pool(X_o) \\
    & V_{attn} = ATTN(X_o) \\
    & V = V_{avg} \oplus V_{max} \oplus V_{attn}
    \end{aligned}
\end{equation}
Lastly, $V$ is fed to linear layers with activations to make a final prediction $\hat y \in [0,1]$, indicating a probability of the input text being hate speech:
\begin{equation}
\small
    \hat y = p(y=1|T) = MLP(V)
\end{equation}
Similarly, the CLP of the input will also go through the same network and get a probability representation, denoted as $\hat{y}^{CLP}$.

\subsection{Loss Function}
\label{Subsec:LossFunction}
Our loss function is designed to consider 1) modules for lowering predictive uncertainty, and 2) fairness for mitigating unintended bias in mixed data.

More specifically, Yao et al.\cite{yao2018rdeepsense} reported scoring rules could overestimate uncertainty (i.e., negative log-likelihood loss -- called NLL loss) or underestimate uncertainty (i.e., mean square error loss -- called MSE loss). They proved and observed that a weighted combination of them could estimate actual uncertainty more accurately in certain scenarios. In general, NLL loss mainly focuses on a macro-level (class-wise) optimization, while MSE loss mainly focuses on a micro-level (instance-wise) optimization. Intuitively they can co-operate with each other. It is worth noting that leveraging MSE loss in classification tasks has recently shown its unique advantage in both theory and experiments~\cite{hui2021evaluation}.
Fairness was also discussed in detail in Section~\ref{subsec:fairness}. We systematically introduce these valuable prior knowledge into the hate speech detection domain by adding gap loss, which measures the difference between an input text and a CLP.

The loss Function $L$ is a weighted sum of three parts: (i) the weighted mean square loss $L_{mse}$; (ii) the weighted negative log-likelihood loss $L_{nll}$; and (iii) the gap loss $L_{gap}$.
\begin{equation}
\small
    \begin{aligned}
    & L_{mse} = \sum_{k=1}^K w_k (y_k-\hat y_k)^2 \\
    & L_{nll} = \sum_{k=1}^K -w_k \log \hat y_k \\
    & L_{gap} = \sum_{k=1}^K |\hat y_k - \hat y'_k| \\
    & L = \gamma L_{mse} + (1 - \gamma) L_{nll} + \lambda L_{gap} + \omega \sum_{l=1}^L ||W^{(l)}||_2^2
    \end{aligned}
\end{equation}
where the last term in $L$ is the L2 regularization, $\gamma , \lambda \in [0,1]$ are tunable hyperparameters. $y_k$ is the label, $\hat y_k$ is the probability for original input and $\hat y'_k$ is the probability for CLP.
As illustrated above, by combining $L_{mse}$ and $L_{nll}$ with a tunable weight $\gamma$, we can approximate the true uncertainty of the model to improve its effectiveness. $\lambda$ being larger will enforce the model to emphasize more on the fairness~\cite{garg2019counterfactual}.
\section{Data}
\label{Sec:Dataset}
\subsection{Dataset}
\label{subsec:originalData}
To conduct experiments, we used datasets from 3 platforms:

\smallskip\noindent\textbf{Twitter:}
 \textbf{Waseem'16\cite{waseem2016hateful}} includes 17,325 tweets which were manually labeled into sexism, racism, offensive, and neither.
    The messages' labels were automatically identified, and the reliability and consistency of labels were manually investigated and verified.
    Offensive speeches do not necessarily lead to hate speech, we filtered out the offensive messages; however, we kept the sexism and racism as hate speech.
\textbf{Davidson'17~\cite{davidson2017automated}} includes 24,783 tweets, consisting of offensive speech, hate speech, and neither. Similarly, we removed offensive speech.
\textbf{Elsherief'18~\cite{elsherief2018hate}} contains only hate speech messages crawled via Twitter Streaming API with specific keywords and hashtags defined by Hatebase\footnote{\url{https://www.hatebase.org/}}. To recognize the anti-hate tweets, which may also contain hate speech terms, the authors cleaned the dataset by using Perspective API\footnote{\url{https://github.com/conversationai/perspectiveapi}} and conducted manual checking during the experiment.

\smallskip\noindent\textbf{Forum:}
\smallskip\noindent\textbf{Degibert'18\cite{degibertetal2018hate}} contains hate and legitimate speeches from forums under possible bias in white supremacy.

\smallskip\noindent\textbf{Wiki:}
\smallskip\noindent\textbf{Wulczyn'17\cite{wulczyn2017ex}}: contains 115k instances where the majority (88\%) of them are legitimate speeches. We leveraged all hate speeches and a sample of legitimate ones to prevent the combined dataset from over imbalanced.

 After retrieving these speeches, we preprocess them and keep speeches longer than three tokens. Overall, our combined dataset consists of 30,762 hate speech messages and 28,693 legitimate messages. 
 Several notes about these datasets need to be clarified and emphasized:
 \squishlist
 \item All datasets are publicly available and have been used in several prior works. 
 \item Unlike previous works~\cite{mou2020swe2,davidson2017automated}, we intentionally chose data across different datasets to enable the best generality of our model. We carefully ensure no single dataset overwhelm others in hate class.
 \item To further justify the data distribution on each platform. We provide statistics in our github repo page. 
 \squishend


\begin{table*}[th]
    \centering
    \caption{Results on the combined dataset. Best baseline: \underline{underlined}, and better results than best baseline: \textbf{bold}.}
    \scalebox{.8}{
    \begin{tabular}{l | c c | c c | c c c}\hline
        \multirow{2}{*}{\textbf{Models}}&    \multicolumn{2}{c|}{\textbf{legit}} &  \multicolumn{2}{c|}{\textbf{hate}} & \multicolumn{3}{c}{\textbf{overall}}\\
        & pre & rec & pre & rec & acc & macF1 & MCC \\\hline
         Davidson'17\cite{davidson2017automated} & $.818_{.048}$ & $.841_{.133}$ & $.862_{.090}$ & $.817_{.083}$ & $.828_{.034}$ & $.826_{.037}$ & $.669_{.064}$\\
         Kim'14\cite{kim2014convolutional} & $.843_{.015}$ & $.852_{.033}$ & $.861_{.023}$ & $.851_{.022}$ & $.852_{.008}$ & $.851_{.008}$ & $.703_{.016}$\\
         Badjatiya'17\cite{badjatiya2017deep} & $.867_{.016}$ & $.865_{.023}$ & $.875_{.017}$ & $.875_{.020}$ & $.870_{.005}$ & $.870_{.005}$ & $.741_{.010}$\\
         Waseem'16\cite{waseem2016hateful} & $.800_{.007}$ & $.918_{.006}$ & $.911_{.006}$ & $.785_{.009}$ & $.849_{.006}$ & $.849_{.007}$ & $.707_{.012}$\\
         Zhang'18\cite{zhang2018detecting} & $.866_{.014}$ & $.840_{.036}$ & $.856_{.025}$ & $.878_{.019}$ & $.860_{.009}$ & $.860_{.009}$ & $.720_{.017}$\\
         Indurthi'19\cite{indurthi2019fermi} & $.872_{.026}$ & $.869_{.034}$ & $.879_{.024}$ & $.879_{.032}$ & $\underline{.874}_{.005}$ & $\underline{.874}_{.005}$ & $\underline{.750}_{.010}$\\
         BERT + LR~\cite{devlin2019bert} & $.853_{.005}$ & $.866_{.005}$ & $.874_{.004}$ & $.861_{.006}$ & $.864_{.004}$ & $.863_{.004}$ & $.727_{.008}$\\
         BERT CLS~\cite{devlin2019bert} & $.863_{.007}$ & $.874_{.011}$ & $.881_{.008}$ & $.870_{.008}$ & $.872_{.003}$ & $.872_{.003}$ & $.744_{.006}$\\
         \hline
         BiQQLSTM & $.909_{.009}$ & $.931_{.011}$ & $.934_{.009}$ & $.913_{.010}$ & $\textbf{.922}_{.005}$ & $\textbf{.922}_{.005}$ & $\textbf{.844}_{.011}$ \\
         BiQQLSTM CLP & $.915_{.010}$ & $.937_{.012}$ & $.940_{.011}$ & $.919_{.010}$ & $\textbf{.927}_{.009}$ & $\textbf{.927}_{.009}$ & $\textbf{.855}_{.018}$ \\
        \hline
    \end{tabular}}
    \label{tab:originalEXP}
    \vspace{-5pt}
\end{table*}

 \subsection{Our Adaptation Method for Data Augmentation}
\label{subsec:augmentation}

We leverage a combination of augmentation methods to push the model performance and robustness to its new limit, namely: \textbf{Charswap}~\cite{mou2020swe2,li2019textbugger,morris2020textattack}; \textbf{Wordnet}~\cite{miller1995wordnet}; \textbf{Embedding}~\cite{wang2015s};  \textbf{Checklist}~\cite{ribeiro2020beyond}; \textbf{Easydata}~\cite{weizou2019eda}; \textbf{NLG}. We make crucial adaptations to the first 5 perturbation methods for reducing data uncertainty. While for the NLG method, we train our task-specific generative model and propose filtering mechanisms for high-quality augmented data. We expect the first 5 methods to provide variations for better robustness while the NLG method for better general performance. 

Perturbation methods might cause label flipping problems, where the context of the sentence would be changed significantly because of certain words' change. We prevented/filtered out changes on certain sensitive words dictionary provided in \cite{elsherief2018hate}. Moreover, negations such as ``no'' and ``not'' were also explicitly prohibited from word addition/deletion methods to prevent augmentations from accidentally reversing the sentence meanings.
For \textbf{Embedding} methods, we applied stricter rules by choosing a threshold 0.8~\cite{morris2020textattack} (i.e., a word was only replaced with another word, which has at least 80\% embedding similarity), and the POS tag matching was required~\cite{rizos2019augment}.

For \textbf{NLG} methods, we finetuned the pre-trained GPT-2 medium separately on hate speeches and legitimate speeches in the training set while keeping the validation set and test set untouched, so resulting in a hate speech generator and a legitimate generator.
We further proposed three methods for controlling NLG's generated content quality:
\begin{enumerate}
    \item We used nucleus sampling~\cite{holtzman2020curious} in decoding to lower the chance of \textbf{repeated words generation}.
    \item \noindent We finetuned another pre-trained BERT model on The Corpus of Linguistic Acceptability (CoLA)~\cite{warstadt2019neural} and used it for removing generated contents which had low \textbf{linguistic acceptability}.
    \item \noindent We believe the \textbf{readability} is also important in generated contents. We used the Flesch readability ease score (FRES)~\cite{flesch1979write} (a lower score means it is harder to read) and kept those speeches with no less than a score of 30 (lower than 30 refers to ``very difficult to read'' or ``extremely difficult to read'') and no larger than 121.22 (the highest possible value in theory).
\end{enumerate}
In addition, we manually checked a randomly sampled 100 hate speeches and 100 legitimate speeches from each augmentation method. 1,199 out of 1,200 samples did not have any mislabeling, confirming our adaptation method's high quality. The only mislabeled sample was created by \textbf{EasyData} method, which deleted context-related words, making the message incomplete. This result gave us a Wilson score confidence interval of [0.9929, 0.9999] under confidence level 99\%.

\subsection{Data Preparation for the Attack Scenario}
\label{subsec:attack}
In essence, the main reason why data augmentation increases model robustness is that injecting mutations in advance (before testing) provides a foreseeable future during training. The model will hopefully learn to capture attack patterns. To simulate the attack scenario where malicious users employ text manipulation/generation methods to generate hate speeches, we used the same data augmentation methods described in Section~\ref{subsec:augmentation} to generate manipulated texts. Unlike the previous data augmentation, in which a source of each method was \emph{the training set}, we used the hate speeches in \emph{the test set} as a source to generate manipulated texts. It means the outcome of the data augmentation and the outcome of this attack scenario would be different. For each attack method, we generated 1,000 hate speeches. To ensure the quality of the generated/manipulated texts, we randomly sampled 100 hate speeches from the 1,000 hate speeches and manually checked them. No mislabeling was found.

\section{Experiment}
\label{Sec:Experiment}

\subsection{Experiment Setting}

\subsubsection{Baselines and our models}
We chose \textbf{8} state-of-the-art baselines to compare against our model: \textbf{Davidson'17\cite{davidson2017automated}, Kim'14\cite{kim2014convolutional}, Badjatiya'17\cite{badjatiya2017deep}, Waseem'16\cite{waseem2016hateful}, Waseem'16\cite{waseem2016hateful}, Zhang'18\cite{zhang2018detecting}, Indurthi'19\cite{indurthi2019fermi}, {BERT + LR}~\cite{devlin2019bert}, and {BERT CLS}~\cite{devlin2019bert}}. We are providing a more detailed description in out github repo due to the space limit.

To ensure a fair comparison, we used the same embedding BERT\_base~\cite{devlin2019bert} for the five baselines: \cite{kim2014convolutional, badjatiya2017deep,zhang2018detecting}, BERT + LR, and BERT CLS. Since the other three baselines originally used either handcrafted features or their own embeddings, which produced better results, so we kept their own design.
It is worth noting that all the above baselines are widely adopted as state-of-the-art ones in recent works.

We run variants of our models: (i) BiQQLSTM (without CLP in the framework and without $L_{gap}$ loss) and (ii) BiQQLSTM CLP (the default model). 

\subsubsection{Train/Validation/Test split}
We conducted a 10-fold cross-validation for our experiments. In each fold, aside from the 10\% hold-out test set, we randomly split the rest into train/validation set with a ratio of 80\%/10\%. Hyper-parameters are tuned on the validation set results. Final results on the 10-fold test sets are reported in average and standard deviations.

\subsubsection{Model Hyperparameters}
\label{sec:parameters}

To improve the model's reproducibility, we report the detailed hyper-parameter values along with their explanations and search space in our github repo.

\subsubsection{Measurements}
We evaluated each model's performance by precision (pre), recall (rec), accuracy (acc), macro F1 score (macF1), and Matthews correlation coefficient (MCC -- a metric especially good for an imbalanced dataset \cite{boughorbel2017optimal}).
The mean and standard deviation (displayed as subscripts in tables in the rest of this paper) are reported.
In Tab.~\ref{tab:attackEXP}, the logistic regression models with the limited-memory BFGS solver did not have randomness inside their frameworks, so we denoted the standard deviation as 0.

\subsection{Experiment results}

\subsubsection{Effectiveness of Our Models}
\label{Subsec:wholedataset}

Tab.~\ref{tab:originalEXP} shows the performance of the eight baselines and our BiQQLSTM without CLP (BiQQLSTM) and BiQQLSTM with CLP (BiQQLSTM CLP) in the combined dataset. Both of our models outperformed the baselines in all overall metrics. In particular, BiQQLSTM CLP achieved 92.2\% accuracy, 0.922 macro F1, and 0.844 MCC, improving up to 5.5\% compared with the best baseline (Indurthi'19). The improvement was statistically significant under a one-tailed t-test (against Indurthi'19). The p-value under accuracy was $7e^{-11}$ (the lower, the more significant).

Another interesting observation is that our BiQQLSTM with CLP performed better than BiQQLSTM (92.7\% vs. 92.2\%). The one-tailed t-test p-value under accuracy is $5e^{-2}$, showing that the difference between the two models is consistent and significant. This result means the fairness (i.e., CLP with $L_{gap}$ loss) prevented unintended social bias and provided data variety into the model as a way of implicit data augmentation.

In addition, we further analyzed how our models and baselines performed for the individual platform's test data, as shown in Tab.~\ref{tab:specificEXP}. Because of the limited space, we only report each model's MCC result. As we described earlier, MCC is a good metric, especially for imbalanced datasets. Again, our models outperformed the baselines in all three datasets, improving more than 15\% on each separate dataset on average.

\begin{table}[t]
    \centering
    \vspace{-5pt}
    \caption{Results on each platform's test set.}
    \scalebox{.8}{
    \begin{tabular}{l | r r r }\hline
        \multirow{2}{*}{\textbf{Models}}&    \textbf{Twitter} & \textbf{Forum} & \textbf{Wiki}\\
         & MCC & MCC & MCC \\\hline
         Davidson'17\cite{davidson2017automated} & $.495_{.046}$ & $.249_{.066}$ & $.607_{.077}$\\
         Kim'14\cite{kim2014convolutional} & $.510_{.040}$ & $.396_{.039}$ & $.664_{.022}$\\
         Badjatiya'17\cite{badjatiya2017deep} &  $.543_{.044}$ & $\underline{.455}_{.036}$ &  $.709_{.014}$\\
         Waseem'16\cite{waseem2016hateful} & $.500_{.045}$ & $.205_{.058}$ &  $.685_{.014}$\\
         Zhang'18\cite{zhang2018detecting} & $.535_{.057}$ & $.420_{.030}$ & $.683_{.023}$\\
         Indurthi'19\cite{indurthi2019fermi} & $\underline{.554}_{.044}$ & $.439_{.062}$ & $\underline{.724}_{.013}$\\
         BERT + LR~\cite{devlin2019bert} & $.471_{.042}$ & $.404_{.031}$ & $.716_{.010}$\\
         BERT CLS~\cite{devlin2019bert} & $.524_{.053}$ & $.441_{.054}$ & $.721_{.013}$ \\\hline
         BiQQLSTM & $\textbf{.681}_{.055}$ & $\textbf{.657}_{.044}$ & $\textbf{.833}_{.012}$ \\
         BiQQLSTM CLP & $\textbf{.687}_{.053}$ & $\textbf{.655}_{.050}$ & $\textbf{.834}_{.013}$ \\
         \hline
    \end{tabular}}
    \label{tab:specificEXP}
\end{table}

For the rest of our experiments, in order to have a detailed look into model variations and to have a fair comparison on the same test set, we report results on one of our 10 folds but with averages and standard deviations on 5 times model rerun with different random seeds. In this way, we prevent the chance of cherry-picking results.

\subsubsection{Effectiveness vs. Efficiency}
\label{Subsec:ablation}

\begin{table}[t]
    \centering
    \caption{Performance and relative training time of our model, and three variants of our model.}
    \scalebox{.8}{
    \begin{tabular}{l | c c c | c}\hline
        \textbf{Models} & acc & macF1 & MCC & Time \\\hline
        BiLSTM CLP & $.919_{.004}$ & $.919_{.004}$ & $.837_{.009}$ & $1.00\times$ \\
        Bi-Quasi-L. CLP & $.912_{.004}$ & $.912_{.004}$ & $.823_{.009}$ & $0.93\times$ \\
        Bi-Quaternion-L. CLP & $.927_{.008}$ & $.927_{.008}$ & $.853_{.016}$ & $1.27\times$\\
        BiQQLSTM CLP & $.927_{.009}$ & $.927_{.009}$ & $.855_{.018}$ & $0.96\times$ \\ \hline
    \end{tabular}}
    \label{tab:Variation}
    \vspace{-5pt}
\end{table}

To understand whether our proposed BiQQLSTM layers helped balance between effectiveness and efficiency, we built 3 additional variants of our model as shown in Tab.~\ref{tab:Variation}.
Given our framework, we replaced BiQQLSTM CLP layers with each of BiLSTM CLP, Bi-Quasi-LSTM CLP, and Bi-Quaternion-LSTM CLP layers. BiQQLSTM CLP and Bi-Quaternion-LSTM CLP performed the best among the 3 variations, but Bi-Quaternion-LSTM CLP took the longest average training time ($1.27\times$ the BiLSTM CLP). Bi-Quasi-LSTM CLP took the shortest training time with a lower performance. Our original BiQQLSTM CLP actually balanced between effectiveness and efficiency, keeping a high-level performance while not downgrading the training time.


Overall, all of our models outperformed all the baselines (refer to Tab.~\ref{tab:originalEXP}), indicating the superiority of our framework and confirming our hypothesis described in Section~\ref{sec:modelDesign}. We also conducted an additional ablation study, and all the components, including uncertainty estimation, positively contributed.


\begin{figure}[ht]
    \centering
    \vspace{-5pt}
    \includegraphics[width=.8\linewidth]{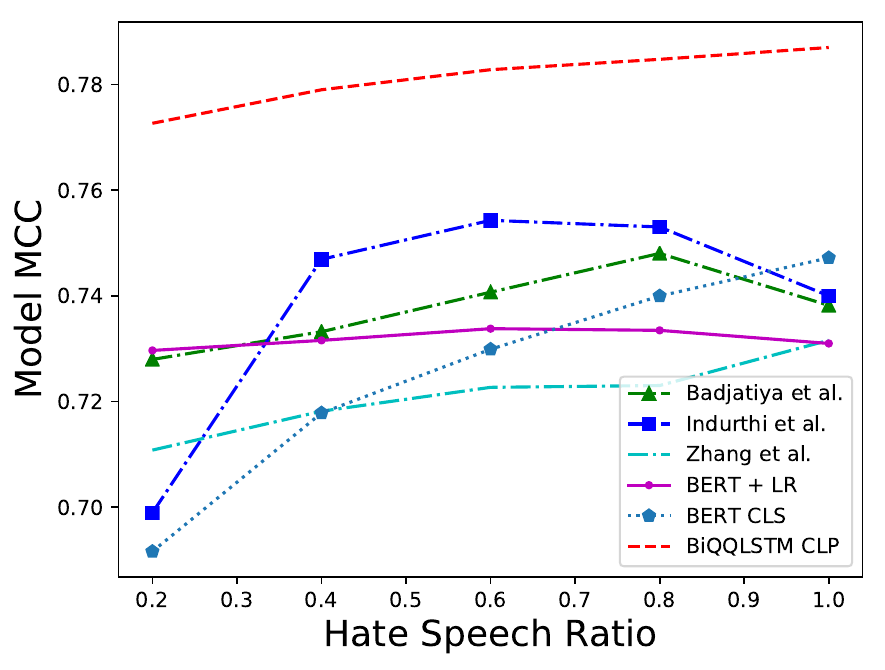}
    \vspace{-5pt}
    \caption{BiQQLSTM CLP VS. Top 5 baselines under avg. MCC of 5 seeds.}
    \label{fig:downsampling}
    \vspace{-10pt}
\end{figure}

\begin{table*}[t]
    \centering
    \caption{Results under various attacks. Best baseline: \underline{underlined}, and better results than best baseline: \textbf{bold}.}
    \scalebox{.8}{
    \begin{tabular}{l | c c c c c c | c}\hline
        \multirow{2}{*}{\textbf{Models}}&    \multicolumn{6}{c|}{\textbf{Precision of Hate Speech Detection}} & ~\\
        & Wordnet & Embedding & Charswap & EasyData & Checklist & NLG & Rank \\\hline
         Davidson'17\cite{davidson2017automated} & $.853_{.007}$ & $.855_{.008}$ & $.851_{.007}$ & $.860_{.007}$ & $.853_{.007}$ & $.879_{.005}$ & 6.7 \\
         Kim'14\cite{kim2014convolutional} & $.854_{.012}$ & $.852_{.017}$ & $.825_{.017}$ & $.829_{.022}$ & $.864_{.019}$ & $.899_{.016}$ & 7.3 \\
         Badjatiya'17\cite{badjatiya2017deep} & $.873_{.015}$ & $.874_{.022}$ & $.837_{.040}$ & $.850_{.032}$ & $.872_{.023}$ & $\underline{.913}_{.022}$ & 5.2 \\
         Waseem'16\cite{waseem2016hateful} & $\underline{.875}_{.000}$ & $\underline{.875}_{.000}$ & $\underline{.874}_{.000}$ & $\underline{.888}_{.000}$ & $\underline{.879}_{.000}$ & $.908_{.000}$ & \underline{3.5} \\
         Zhang'18\cite{zhang2018detecting} & $.855_{.030}$ & $.853_{.031}$ & $.828_{.028}$ & $.828_{.033}$ & $.848_{.040}$ & $.882_{.036}$ & 7.5 \\
         Indurthi'19\cite{indurthi2019fermi} & $.821_{.053}$ & $.826_{.053}$ & $.813_{.050}$ & $.805_{.050}$ & $.823_{.053}$ & $.839_{.050}$ & 10.8 \\
         BERT + LR~\cite{devlin2019bert} & $.835_{.000}$ & $.837_{.000}$ & $.817_{.000}$ & $.833_{.000}$ & $.846_{.000}$ & $.880_{.000}$ & 8.5 \\
         BERT CLS~\cite{devlin2019bert} & $.835_{.008}$ & $.837_{.012}$ & $.811_{.012}$ & $.833_{.007}$ & $.846_{.012}$ & $.877_{.017}$ & 9.2 \\\hline
         BiQQLSTM CLP & $\textbf{.877}_{.007}$ & $\textbf{.883}_{.008}$ & $.870_{.007}$ & $.885_{.005}$ & $\textbf{.894}_{.008}$ & $\textbf{.933}_{.006}$ & \textbf{3.2}\\
         BiQQLSTM CLP NLG+Checklist+Embedding & $\textbf{.879}_{.012}$ & $\textbf{.884}_{.013}$ & $.862_{.012}$ & $\textbf{.888}_{.011}$ & $\textbf{.910}_{.014}$ & $\textbf{.943}_{.010}$ & \textbf{2.0}\\
         BiQQLSTM CLP FullAug & $\textbf{.881}_{.009}$ & $\textbf{.886}_{.009}$ & $\textbf{.876}_{.009}$ & $\textbf{.889}_{.008}$ & $\textbf{.906}_{.009}$ & $\textbf{.942}_{.005}$ & \textbf{1.3} \\ \hline
    \end{tabular}}
    \label{tab:attackEXP}
    \vspace{-5pt}
\end{table*}

\subsubsection{Varying Hate Speech Ratio}
\label{subsec:downsampling}
Considering real-world cases where the actual amount of legitimate speech may overwhelm the hate speeches, we tested the effectiveness of our model under various imbalanced class ratios by downsampling the amount of hate speech data in the training set. We show the result of the top-5 baselines vs. our model (BiQQLSTM CLP) in Fig.~\ref{fig:downsampling}. As the classes are now imbalanced, we report the average results under MCC for 5 different seeds. We observe similar patterns as reported in Section~\ref{Subsec:wholedataset}, where our BiQQLSTM CLP(in red on top of the figure) consistently reaches the best performance, confirming the effectiveness of our model regardless of a hate speech ratio.

\subsubsection{Effectiveness of Data Augmentation}
\label{Subsec:augmentVariation}

\begin{figure}[t]
    \centering
    \includegraphics[width=.8\linewidth]{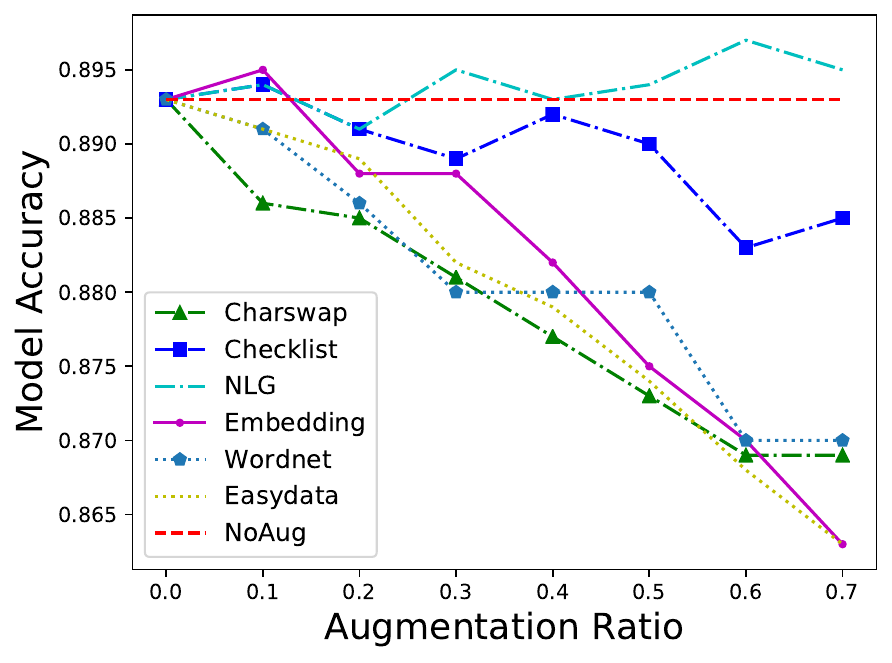}
    \vspace{-5pt}
    \caption{Our BiQQLSTM CLP's performance with each augmentation method.}
    \vspace{-10pt}
    \label{fig:augRatio}
\end{figure}

We tested how each data augmentation method with our adaptation would be helpful by adding its generated data into the training set and re-built our model. 
We used BiQQLSTM CLP as our default framework and fed the generated data and the training set (i.e., 80\% of the combined dataset) into the framework. Fig.~\ref{fig:augRatio} shows how the effectiveness of our model changed as we increased the amount of augmented data obtained from each method. The x-axis represents the augmentation ratio compared with the original training set. 
The y-axis represents the model's performance (Accuracy) on the test set. We draw one horizontal line, which represents BiQQLSTM CLP without any augmentation (NoAug), to compare whether augmentation is helpful. 
Overall, \textbf{NLG} method kept steady performance across different ratios. \textbf{Checklist} and \textbf{Embedding} were also helpful if we only add 0.1 (10\%) augmentation ratio. However, \textbf{Charswap} and \textbf{Easydata} were not helpful because \textbf{Charswap} might create more misspelling, and \textbf{Easydata}'s randomness might add noise into the model by losing the context.


\subsubsection{Effectiveness of Data Augmentation under Attack Scenario}
\label{Subsec:robustness}


As we described in Section~\ref{subsec:attack} regarding how we prepared data for the attack scenario, we created manipulated data for the attack scenario from the testing set to test whether our model with data augmentation (learned from the training set) is effective against the attacks. As the original training data and testing data are separate, there is no information leak between them and thus the attack scenario is non-trivial.

To understand whether the previously mentioned data augmentation methods further improve our model's robustness under the attack, we made two variants of our model called BiQQLSTM CLP NLG+Checklist+Embedding and BiQQLSTM CLP FullAug. 
In the two variants, the best augmentation ratios were selected from the previous experiment in Fig.~\ref{fig:augRatio}.

Tab.~\ref{tab:attackEXP} shows experiment results and average rank (lower is better) on each column/attack of our models with or without data augmentation and the baselines. Among the baselines, surprisingly, \cite{indurthi2019fermi} and BERT CLS were the most vulnerable under the attack, although they were the among best baselines in the previous experiment (refer to Tab.~\ref{tab:originalEXP}). It means a model's effectiveness may not guarantee its robustness. It also means our framework is well designed for both effectiveness and robustness. Overall, all of our models were the most robust compared with the baselines. We see three interesting observations: (1) Our model without data augmentation (BiQQLSTM CLP) was still more robust than the baselines. It confirms the superiority of our framework compared with the baselines in terms of both effectiveness and robustness. (2) All of our models performed very well in Embedding and NLG attacks, which are more advanced on machine-generated attacks. (3) Adding texts generated by corresponding data augmentation methods into BiQQLSTM CLP would further improve its robustness against the same attack method.

\subsection{Effectiveness of our Filtering Methods for NLG Augment.}
\label{sec:analysis}
\begin{figure}[ht]
\centering
\vspace{-5pt}
\subfloat[Before filtering\label{fig:generatedReadability}]{\includegraphics[width=0.75\linewidth]{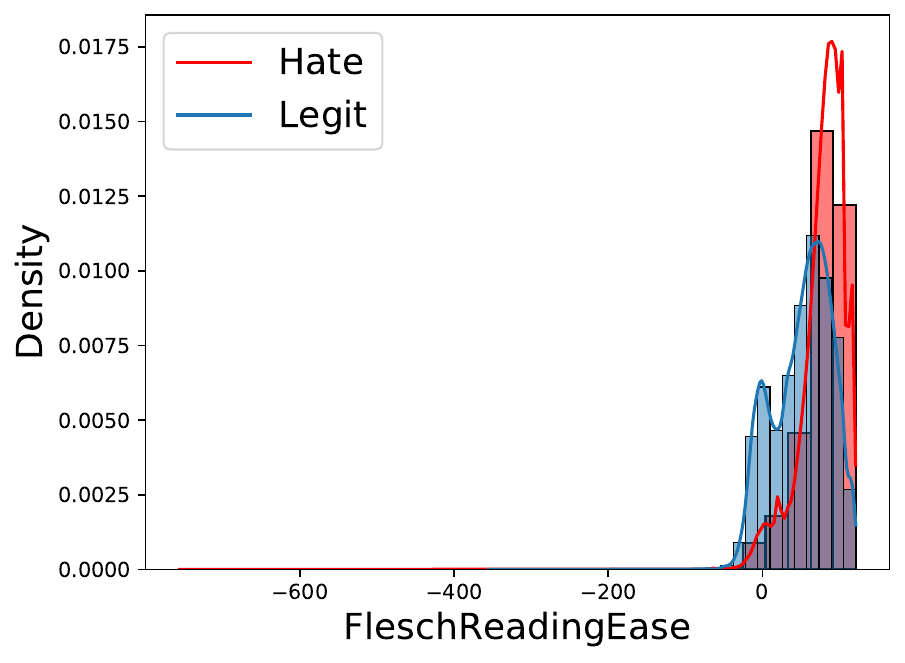}}\\
\subfloat[After filtering\label{fig:filteredReadability}]{\includegraphics[width=0.73\linewidth]{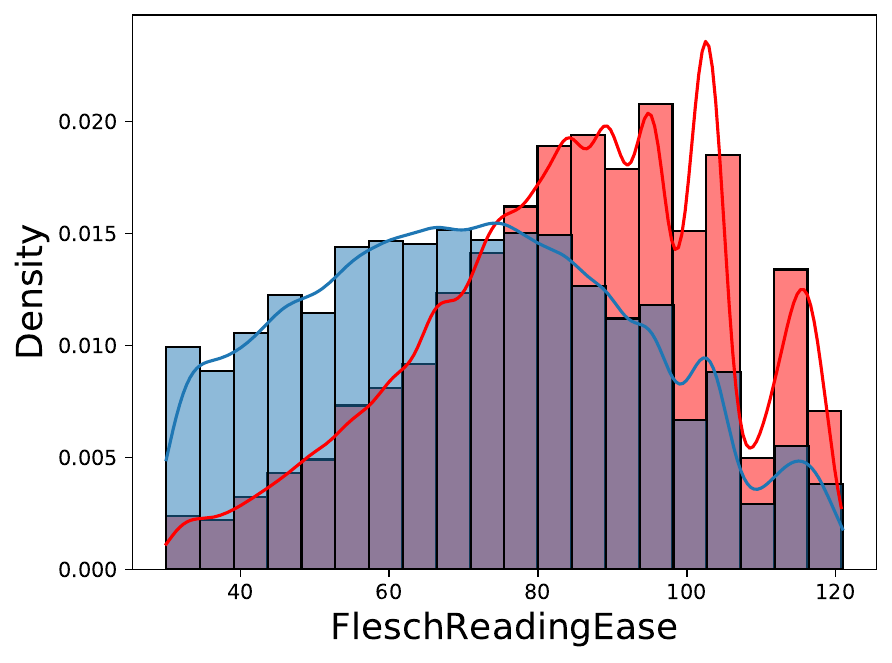}}
\caption{NLG augmented data readability distribution before \& after filtering.}
\label{fig:readability}
\vspace{-10pt}
\end{figure}

To confirm the effectiveness of our proposed filtering methods to improve NLG augmentation quality, we show the readability distribution before and after applying filtering rules in Fig.~\ref{fig:readability}. We still observed a long tail on the left side before filtering, although we already removed generated data that had less than -800 score (very hard-to-read contents) in Fig.~\ref{fig:generatedReadability}. Filtering out low-quality data essentially helped to keep only high-quality augmented data, as shown in Fig.~\ref{fig:filteredReadability}. 
%




\section{Conclusion}
\label{Sec:Conclusion}
In this paper, we have proposed a novel data-augmented and fairness-aware BiQQLSTM framework for improving model performance, robustness, and fairness in hate speech detection. Our model has outperformed all baselines, improving up to 5.5\% under no attack scenario and up to 3.1\% under the attack scenario compared with the best baseline in each scenario. Our model managed to achieve both effectiveness and robustness according to our experiment results.

\section*{Acknowledgment}
\noindent This work was supported in part by NSF grant CNS-1755536. 

\bibliographystyle{./bibliography/IEEEtran}
\bibliography{./bibliography/IEEEabrv,./bibliography/IEEEexample}

\begin{thebibliography}{10}
\providecommand{\url}[1]{#1}
\csname url@samestyle\endcsname
\providecommand{\newblock}{\relax}
\providecommand{\bibinfo}[2]{#2}
\providecommand{\BIBentrySTDinterwordspacing}{\spaceskip=0pt\relax}
\providecommand{\BIBentryALTinterwordstretchfactor}{4}
\providecommand{\BIBentryALTinterwordspacing}{\spaceskip=\fontdimen2\font plus
\BIBentryALTinterwordstretchfactor\fontdimen3\font minus
  \fontdimen4\font\relax}
\providecommand{\BIBforeignlanguage}[2]{{%
\expandafter\ifx\csname l@#1\endcsname\relax
\typeout{** WARNING: IEEEtran.bst: No hyphenation pattern has been}%
\typeout{** loaded for the language `#1'. Using the pattern for}%
\typeout{** the default language instead.}%
\else
\language=\csname l@#1\endcsname
\fi
#2}}
\providecommand{\BIBdecl}{\relax}
\BIBdecl

\bibitem{laub2019hate}
Z.~Laub, ``Hate speech on social media: Global comparisons,'' \emph{Council on
  Foreign Relations}, vol.~7, 2019.

\bibitem{nobata2016abusive}
\BIBentryALTinterwordspacing
C.~Nobata, J.~Tetreault, A.~Thomas, Y.~Mehdad, and Y.~Chang, ``Abusive language
  detection in online user content,'' in \emph{WWW}, 2016. [Online]. Available:
  \url{https://doi.org/10.1145/2872427.2883062}
\BIBentrySTDinterwordspacing

\bibitem{djuric2015hate}
\BIBentryALTinterwordspacing
N.~Djuric, J.~Zhou, R.~Morris, M.~Grbovic, V.~Radosavljevic, and
  N.~Bhamidipati, ``Hate speech detection with comment embeddings,'' in
  \emph{WWW}, 2015. [Online]. Available:
  \url{https://doi.org/10.1145/2740908.2742760}
\BIBentrySTDinterwordspacing

\bibitem{badjatiya2019stereotypical}
\BIBentryALTinterwordspacing
P.~Badjatiya, M.~Gupta, and V.~Varma, ``Stereotypical bias removal for hate
  speech detection task using knowledge-based generalizations,'' in \emph{WWW},
  2019. [Online]. Available: \url{https://doi.org/10.1145/3308558.3313504}
\BIBentrySTDinterwordspacing

\bibitem{liu2019fuzzy}
\BIBentryALTinterwordspacing
H.~Liu, P.~Burnap, W.~Alorainy, and M.~L. Williams, ``Fuzzy multi-task learning
  for hate speech type identification,'' in \emph{WWW}, 2019. [Online].
  Available: \url{https://doi.org/10.1145/3308558.3313546}
\BIBentrySTDinterwordspacing

\bibitem{chowdhury2019arhnet}
\BIBentryALTinterwordspacing
A.~G. Chowdhury, A.~Didolkar, R.~Sawhney, and R.~Shah, ``Arhnet-leveraging
  community interaction for detection of religious hate speech in arabic,'' in
  \emph{ACL Student Research Workshop}, 2019. [Online]. Available:
  \url{https://doi.org/10.18653/v1/p19-2038}
\BIBentrySTDinterwordspacing

\bibitem{zhang2018detecting}
\BIBentryALTinterwordspacing
Z.~Zhang, D.~Robinson, and J.~Tepper, ``Detecting hate speech on twitter using
  a convolution-gru based deep neural network,'' in \emph{European semantic web
  conference}, 2018. [Online]. Available:
  \url{https://doi.org/10.1007/978-3-319-93417-4_48}
\BIBentrySTDinterwordspacing

\bibitem{degibertetal2018hate}
\BIBentryALTinterwordspacing
O.~de~Gibert, N.~Perez, A.~Garc{\'\i}a-Pablos, and M.~Cuadros, ``Hate speech
  dataset from a white supremacy forum,'' in \emph{2nd Workshop on Abusive
  Language Online ({ALW}2)}.\hskip 1em plus 0.5em minus 0.4em\relax Brussels,
  Belgium: Association for Computational Linguistics, Oct. 2018, pp. 11--20.
  [Online]. Available: \url{https://www.aclweb.org/anthology/W18-5102}
\BIBentrySTDinterwordspacing

\bibitem{wulczyn2017ex}
\BIBentryALTinterwordspacing
E.~Wulczyn, N.~Thain, and L.~Dixon, ``Ex machina: Personal attacks seen at
  scale,'' in \emph{WWW}, 2017, pp. 1391--1399. [Online]. Available:
  \url{https://doi.org/10.1145/3038912.3052591}
\BIBentrySTDinterwordspacing

\bibitem{alrehili2019automatic}
\BIBentryALTinterwordspacing
A.~Alrehili, ``Automatic hate speech detection on social media: A brief
  survey,'' in \emph{AICCSA}, 2019. [Online]. Available:
  \url{https://doi.org/10.1109/aiccsa47632.2019.9035228}
\BIBentrySTDinterwordspacing

\bibitem{ousidhoum2019multilingual}
\BIBentryALTinterwordspacing
N.~Ousidhoum, Z.~Lin, H.~Zhang, Y.~Song, and D.-Y. Yeung, ``Multilingual and
  multi-aspect hate speech analysis,'' in \emph{EMNLP-IJCNLP}.\hskip 1em plus
  0.5em minus 0.4em\relax Hong Kong, China: Association for Computational
  Linguistics, Nov. 2019, pp. 4675--4684. [Online]. Available:
  \url{https://www.aclweb.org/anthology/D19-1474}
\BIBentrySTDinterwordspacing

\bibitem{davidson2017automated}
T.~Davidson, D.~Warmsley, M.~Macy, and I.~Weber, ``Automated hate speech
  detection and the problem of offensive language,'' in \emph{Proceedings of
  the International AAAI Conference on Web and Social Media}, vol.~11, no.~1,
  2017.

\bibitem{waseem2016hateful}
\BIBentryALTinterwordspacing
Z.~Waseem and D.~Hovy, ``Hateful symbols or hateful people? predictive features
  for hate speech detection on twitter,'' in \emph{NAACL student research
  workshop}, 2016, pp. 88--93. [Online]. Available:
  \url{https://doi.org/10.18653/v1/n16-2013}
\BIBentrySTDinterwordspacing

\bibitem{elsherief2018hate}
M.~ElSherief, V.~Kulkarni, D.~Nguyen, W.~Y. Wang, and E.~Belding, ``Hate lingo:
  A target-based linguistic analysis of hate speech in social media,'' in
  \emph{ICWSM}, vol.~12, no.~1, 2018.

\bibitem{garg2019counterfactual}
\BIBentryALTinterwordspacing
S.~Garg, V.~Perot, N.~Limtiaco, A.~Taly, E.~H. Chi, and A.~Beutel,
  ``Counterfactual fairness in text classification through robustness,'' in
  \emph{AAAI/ACM Conference on AI, Ethics, and Society}, 2019, pp. 219--226.
  [Online]. Available: \url{https://doi.org/10.1145/3306618.3317950}
\BIBentrySTDinterwordspacing

\bibitem{warner2012detecting}
W.~Warner and J.~Hirschberg, ``Detecting hate speech on the world wide web,''
  in \emph{Second workshop on language in social media}, 2012.

\bibitem{waseem2016you}
\BIBentryALTinterwordspacing
Z.~Waseem, ``Are you a racist or am i seeing things? annotator influence on
  hate speech detection on twitter,'' in \emph{Workshop on NLP and
  computational social science}, 2016, pp. 138--142. [Online]. Available:
  \url{https://doi.org/10.18653/v1/w16-5618}
\BIBentrySTDinterwordspacing

\bibitem{arango2019hate}
\BIBentryALTinterwordspacing
A.~Arango, J.~P{\'e}rez, and B.~Poblete, ``Hate speech detection is not as easy
  as you may think: A closer look at model validation,'' in \emph{SIGIR}, 2019,
  pp. 45--54. [Online]. Available:
  \url{https://doi.org/10.1016/j.is.2020.101584}
\BIBentrySTDinterwordspacing

\bibitem{badjatiya2017deep}
\BIBentryALTinterwordspacing
P.~Badjatiya, S.~Gupta, M.~Gupta, and V.~Varma, ``Deep learning for hate speech
  detection in tweets,'' in \emph{International Conference on World Wide Web
  Companion}, 2017. [Online]. Available:
  \url{https://doi.org/10.1145/3041021.3054223}
\BIBentrySTDinterwordspacing

\bibitem{gamback2017using}
\BIBentryALTinterwordspacing
B.~Gamb{\"a}ck and U.~K. Sikdar, ``Using convolutional neural networks to
  classify hate-speech,'' in \emph{Workshop on abusive language online}, 2017,
  pp. 85--90. [Online]. Available: \url{https://doi.org/10.18653/v1/w17-3013}
\BIBentrySTDinterwordspacing

\bibitem{mou2020swe2}
\BIBentryALTinterwordspacing
G.~Mou, P.~Ye, and K.~Lee, ``Swe2: Subword enriched and significant word
  emphasized framework for hate speech detection,'' in \emph{CIKM}, 2020, pp.
  1145--1154. [Online]. Available:
  \url{https://doi.org/10.1145/3340531.3411990}
\BIBentrySTDinterwordspacing

\bibitem{sun2020adv}
L.~Sun, K.~Hashimoto, W.~Yin, A.~Asai, J.~Li, P.~Yu, and C.~Xiong, ``Adv-bert:
  Bert is not robust on misspellings! generating nature adversarial samples on
  bert,'' \emph{arXiv preprint arXiv:2003.04985}, 2020.

\bibitem{garg2020bae}
\BIBentryALTinterwordspacing
S.~Garg and G.~Ramakrishnan, ``{BAE}: {BERT}-based adversarial examples for
  text classification,'' in \emph{EMNLP}, 2020. [Online]. Available:
  \url{https://doi.org/10.18653/v1/2020.emnlp-main.498}
\BIBentrySTDinterwordspacing

\bibitem{li2020bert}
\BIBentryALTinterwordspacing
L.~Li, R.~Ma, Q.~Guo, X.~Xue, and X.~Qiu, ``Bert-attack: Adversarial attack
  against bert using bert,'' in \emph{EMNLP}, 2020. [Online]. Available:
  \url{https://doi.org/10.18653/v1/2020.emnlp-main.500}
\BIBentrySTDinterwordspacing

\bibitem{li2020contextualized}
D.~Li, Y.~Zhang, H.~Peng, L.~Chen, C.~Brockett, M.-T. Sun, and B.~Dolan,
  ``Contextualized perturbation for textual adversarial attack,'' \emph{arXiv
  preprint arXiv:2009.07502}, 2020.

\bibitem{mathew2019thou}
B.~Mathew, P.~Saha, H.~Tharad, S.~Rajgaria, P.~Singhania, S.~K. Maity,
  P.~Goyal, and A.~Mukherjee, ``Thou shalt not hate: Countering online hate
  speech,'' in \emph{ICWSM}, vol.~13, 2019, pp. 369--380.

\bibitem{chung2019conan}
\BIBentryALTinterwordspacing
Y.-L. Chung, E.~Kuzmenko, S.~S. Tekiroglu, and M.~Guerini, ``{CONAN} -
  {CO}unter {NA}rratives through nichesourcing: a multilingual dataset of
  responses to fight online hate speech,'' in \emph{ACL}.\hskip 1em plus 0.5em
  minus 0.4em\relax Florence, Italy: Association for Computational Linguistics,
  Jul. 2019, pp. 2819--2829. [Online]. Available:
  \url{https://www.aclweb.org/anthology/P19-1271}
\BIBentrySTDinterwordspacing

\bibitem{tran2020habertor}
\BIBentryALTinterwordspacing
T.~Tran, Y.~Hu, C.~Hu, K.~Yen, F.~Tan, K.~Lee, and S.~Park, ``Habertor: An
  efficient and effective deep hatespeech detector,'' in \emph{EMNLP}, 2020.
  [Online]. Available: \url{https://doi.org/10.18653/v1/2020.emnlp-main.606}
\BIBentrySTDinterwordspacing

\bibitem{miller1995wordnet}
G.~A. Miller, ``Wordnet: a lexical database for english,'' \emph{Communications
  of the ACM}, vol.~38, no.~11, pp. 39--41, 1995.

\bibitem{alzantot2018generating}
\BIBentryALTinterwordspacing
M.~Alzantot, Y.~Sharma, A.~Elgohary, B.-J. Ho, M.~Srivastava, and K.-W. Chang,
  ``Generating natural language adversarial examples,'' in \emph{EMNLP}, 2018.
  [Online]. Available: \url{https://doi.org/10.18653/v1/d18-1316}
\BIBentrySTDinterwordspacing

\bibitem{ribeiro2020beyond}
\BIBentryALTinterwordspacing
M.~T. Ribeiro, T.~Wu, C.~Guestrin, and S.~Singh, ``Beyond accuracy: Behavioral
  testing of nlp models with checklist,'' in \emph{ACL}, 2020. [Online].
  Available: \url{https://doi.org/10.18653/v1/2020.acl-main.442}
\BIBentrySTDinterwordspacing

\bibitem{jiao2020tinybert}
\BIBentryALTinterwordspacing
X.~Jiao, Y.~Yin, L.~Shang, X.~Jiang, X.~Chen, L.~Li, F.~Wang, and Q.~Liu,
  ``Tinybert: Distilling {BERT} for natural language understanding,'' in
  \emph{EMNLP}, T.~Cohn, Y.~He, and Y.~Liu, Eds.\hskip 1em plus 0.5em minus
  0.4em\relax Association for Computational Linguistics, 2020, pp. 4163--4174.
  [Online]. Available:
  \url{https://doi.org/10.18653/v1/2020.findings-emnlp.372}
\BIBentrySTDinterwordspacing

\bibitem{wu2019conditional}
X.~Wu, S.~Lv, L.~Zang, J.~Han, and S.~Hu, ``Conditional bert contextual
  augmentation,'' in \emph{ICCS}.\hskip 1em plus 0.5em minus 0.4em\relax
  Springer, 2019, pp. 84--95.

\bibitem{xie2020unsupervised}
\BIBentryALTinterwordspacing
Q.~Xie, Z.~Dai, E.~H. Hovy, T.~Luong, and Q.~Le, ``Unsupervised data
  augmentation for consistency training,'' in \emph{NeurIPS}, H.~Larochelle,
  M.~Ranzato, R.~Hadsell, M.~Balcan, and H.~Lin, Eds., 2020. [Online].
  Available:
  \url{https://proceedings.neurips.cc/paper/2020/hash/44feb0096faa8326192570788b38c1d1-Abstract.html}
\BIBentrySTDinterwordspacing

\bibitem{luque2019atalaya}
\BIBentryALTinterwordspacing
F.~M. Luque, ``Atalaya at {TASS} 2019: Data augmentation and robust embeddings
  for sentiment analysis,'' in \emph{Proceedings of the Iberian Languages
  Evaluation Forum co-located with 35th Conference of the Spanish Society for
  Natural Language Processing, IberLEF@SEPLN 2019, Bilbao, Spain, September
  24th, 2019}, ser. {CEUR} Workshop Proceedings, M.~{\'{A}}.~G. Cumbreras,
  J.~Gonzalo, E.~M. C{\'{a}}mara, R.~Mart{\'{\i}}nez{-}Unanue, P.~Rosso,
  J.~Carrillo{-}de{-}Albornoz, S.~Montalvo, L.~Chiruzzo, S.~Collovini,
  Y.~Guti{\'{e}}rrez, S.~M.~J. Zafra, M.~Krallinger,
  M.~Montes{-}y{-}G{\'{o}}mez, R.~Ortega{-}Bueno, and A.~Ros{\'{a}}, Eds., vol.
  2421.\hskip 1em plus 0.5em minus 0.4em\relax CEUR-WS.org, 2019, pp. 561--570.
  [Online]. Available: \url{http://ceur-ws.org/Vol-2421/TASS\_paper\_1.pdf}
\BIBentrySTDinterwordspacing

\bibitem{li2019textbugger}
\BIBentryALTinterwordspacing
J.~Li, S.~Ji, T.~Du, B.~Li, and T.~Wang, ``Textbugger: Generating adversarial
  text against real-world applications,'' in \emph{NDSS}, 2019. [Online].
  Available: \url{https://doi.org/10.14722/ndss.2019.23138}
\BIBentrySTDinterwordspacing

\bibitem{pruthi2019combating}
\BIBentryALTinterwordspacing
D.~Pruthi, B.~Dhingra, and Z.~C. Lipton, ``Combating adversarial misspellings
  with robust word recognition,'' in \emph{ACL}, 2019. [Online]. Available:
  \url{https://doi.org/10.18653/v1/p19-1561}
\BIBentrySTDinterwordspacing

\bibitem{shen2020simple}
D.~Shen, M.~Zheng, Y.~Shen, Y.~Qu, and W.~Chen, ``A simple but tough-to-beat
  data augmentation approach for natural language understanding and
  generation,'' \emph{arXiv preprint arXiv:2009.13818}, 2020.

\bibitem{weizou2019eda}
\BIBentryALTinterwordspacing
J.~Wei and K.~Zou, ``{EDA}: Easy data augmentation techniques for boosting
  performance on text classification tasks,'' in \emph{EMNLP-IJCNLP}.\hskip 1em
  plus 0.5em minus 0.4em\relax Hong Kong, China: Association for Computational
  Linguistics, Nov. 2019, pp. 6382--6388. [Online]. Available:
  \url{https://www.aclweb.org/anthology/D19-1670}
\BIBentrySTDinterwordspacing

\bibitem{coulombe2018text}
\BIBentryALTinterwordspacing
C.~Coulombe, ``Text data augmentation made simple by leveraging {NLP} cloud
  apis,'' \emph{CoRR}, vol. abs/1812.04718, 2018. [Online]. Available:
  \url{http://arxiv.org/abs/1812.04718}
\BIBentrySTDinterwordspacing

\bibitem{rizos2019augment}
\BIBentryALTinterwordspacing
G.~Rizos, K.~Hemker, and B.~Schuller, ``Augment to prevent: short-text data
  augmentation in deep learning for hate-speech classification,'' in
  \emph{CIKM}, 2019, pp. 991--1000. [Online]. Available:
  \url{https://doi.org/10.1145/3357384.3358040}
\BIBentrySTDinterwordspacing

\bibitem{yi2021reweighting}
\BIBentryALTinterwordspacing
M.~Yi, L.~Hou, L.~Shang, X.~Jiang, Q.~Liu, and Z.-M. Ma, ``Reweighting
  augmented samples by minimizing the maximal expected loss,'' in \emph{ICLR},
  2021. [Online]. Available: \url{https://openreview.net/forum?id=9G5MIc-goqB}
\BIBentrySTDinterwordspacing

\bibitem{wullach2020towards}
\BIBentryALTinterwordspacing
T.~{Wullach}, A.~{Adler}, and E.~M. {Minkov}, ``Towards hate speech detection
  at large via deep generative modeling,'' \emph{IEEE Internet Computing}, pp.
  1--1, 2020. [Online]. Available:
  \url{https://doi.org/10.1109/mic.2020.3033161}
\BIBentrySTDinterwordspacing

\bibitem{radford2019language}
A.~Radford, J.~Wu, R.~Child, D.~Luan, D.~Amodei, and I.~Sutskever, ``Language
  models are unsupervised multitask learners,'' \emph{OpenAI blog}, vol.~1,
  no.~8, p.~9, 2019.

\bibitem{cao2020hategan}
\BIBentryALTinterwordspacing
R.~Cao and R.~K.-W. Lee, ``Hategan: Adversarial generative-based data
  augmentation for hate speech detection,'' in \emph{COLING}, 2020, pp.
  6327--6338. [Online]. Available:
  \url{https://doi.org/10.18653/v1/2020.coling-main.557}
\BIBentrySTDinterwordspacing

\bibitem{anaby2020not}
A.~Anaby-Tavor, B.~Carmeli, E.~Goldbraich, A.~Kantor, G.~Kour, S.~Shlomov,
  N.~Tepper, and N.~Zwerdling, ``Do not have enough data? deep learning to the
  rescue!'' in \emph{AAAI}, vol.~34, no.~05, 2020, pp. 7383--7390.

\bibitem{kumar2020data}
V.~Kumar, A.~Choudhary, and E.~Cho, ``Data augmentation using pre-trained
  transformer models,'' \emph{arXiv preprint arXiv:2003.02245}, 2020.

\bibitem{wang2015s}
W.~Y. Wang and D.~Yang, ``That’s so annoying!!!: A lexical and frame-semantic
  embedding based data augmentation approach to automatic categorization of
  annoying behaviors using\# petpeeve tweets,'' in \emph{EMNLP}, 2015, pp.
  2557--2563.

\bibitem{frezal2019fairness}
\BIBentryALTinterwordspacing
S.~Frezal and L.~Barry, ``Fairness in uncertainty: Some limits and
  misinterpretations of actuarial fairness,'' \emph{Journal of Business
  Ethics}, pp. 1--10, 2019. [Online]. Available:
  \url{https://doi.org/10.1007/s10551-019-04171-2}
\BIBentrySTDinterwordspacing

\bibitem{wang2020fairness}
L.~Wang and T.~Joachims, ``Fairness and diversity for rankings in two-sided
  markets,'' \emph{arXiv preprint arXiv:2010.01470}, 2020.

\bibitem{patro2020incremental}
\BIBentryALTinterwordspacing
G.~K. Patro, A.~Chakraborty, N.~Ganguly, and K.~Gummadi, ``Incremental fairness
  in two-sided market platforms: On smoothly updating recommendations,'' in
  \emph{AAAI Conference on Artificial Intelligence}, 2020, pp. 181--188.
  [Online]. Available: \url{https://doi.org/10.1609/aaai.v34i01.5349}
\BIBentrySTDinterwordspacing

\bibitem{nissim2020fair}
\BIBentryALTinterwordspacing
M.~Nissim, R.~van Noord, and R.~van~der Goot, ``Fair is better than
  sensational: Man is to doctor as woman is to doctor,'' \emph{Computational
  Linguistics}, 2020. [Online]. Available:
  \url{https://doi.org/10.1162/coli_a_00379}
\BIBentrySTDinterwordspacing

\bibitem{hardt2016equality}
M.~Hardt, E.~Price, and N.~Srebro, ``Equality of opportunity in supervised
  learning,'' \emph{NeurIPS}, vol.~29, pp. 3315--3323, 2016.

\bibitem{zhang2018equality}
J.~Zhang and E.~Bareinboim, ``Equality of opportunity in classification: A
  causal approach,'' in \emph{NeurIPS}, 2018, pp. 3671--3681.

\bibitem{beutel2017data}
A.~Beutel, J.~Chen, Z.~Zhao, and E.~H. Chi, ``Data decisions and theoretical
  implications when adversarially learning fair representations,'' in
  \emph{Workshop on Fairness, Accountability, and Transparency in Machine
  Learning (FAT/ML 2017)}, 2017.

\bibitem{zhang2018mitigating}
\BIBentryALTinterwordspacing
B.~H. Zhang, B.~Lemoine, and M.~Mitchell, ``Mitigating unwanted biases with
  adversarial learning,'' in \emph{AAAI/ACM Conference on AI, Ethics, and
  Society}, 2018, pp. 335--340. [Online]. Available:
  \url{https://doi.org/10.1145/3278721.3278779}
\BIBentrySTDinterwordspacing

\bibitem{celis2019classification}
\BIBentryALTinterwordspacing
L.~E. Celis, L.~Huang, V.~Keswani, and N.~K. Vishnoi, ``Classification with
  fairness constraints: A meta-algorithm with provable guarantees,'' in
  \emph{Conference on Fairness, Accountability, and Transparency}, 2019, pp.
  319--328. [Online]. Available: \url{https://doi.org/10.1145/3287560.3287586}
\BIBentrySTDinterwordspacing

\bibitem{dixon2018measuring}
\BIBentryALTinterwordspacing
L.~Dixon, J.~Li, J.~Sorensen, N.~Thain, and L.~Vasserman, ``Measuring and
  mitigating unintended bias in text classification,'' in \emph{AAAI/ACM
  Conference on AI, Ethics, and Society}, 2018, pp. 67--73. [Online].
  Available: \url{https://doi.org/10.1145/3278721.3278729}
\BIBentrySTDinterwordspacing

\bibitem{lakshminarayanan2017simple}
B.~Lakshminarayanan, A.~Pritzel, and C.~Blundell, ``Simple and scalable
  predictive uncertainty estimation using deep ensembles,'' in \emph{NeurIPS},
  2017, pp. 6402--6413.

\bibitem{gal2016dropout}
Y.~Gal and Z.~Ghahramani, ``Dropout as a bayesian approximation: Representing
  model uncertainty in deep learning,'' in \emph{ICML}, 2016, pp. 1050--1059.

\bibitem{loquercio2020general}
\BIBentryALTinterwordspacing
A.~Loquercio, M.~Segu, and D.~Scaramuzza, ``A general framework for uncertainty
  estimation in deep learning,'' \emph{IEEE Robotics and Automation Letters},
  vol.~5, no.~2, pp. 3153--3160, 2020. [Online]. Available:
  \url{https://doi.org/10.1109/lra.2020.2974682}
\BIBentrySTDinterwordspacing

\bibitem{borsuk2004bayesian}
\BIBentryALTinterwordspacing
M.~E. Borsuk, C.~A. Stow, and K.~H. Reckhow, ``A bayesian network of
  eutrophication models for synthesis, prediction, and uncertainty analysis,''
  \emph{Ecological Modelling}, vol. 173, no. 2-3, pp. 219--239, 2004. [Online].
  Available: \url{https://doi.org/10.1016/j.ecolmodel.2003.08.020}
\BIBentrySTDinterwordspacing

\bibitem{gaudet2018deep}
\BIBentryALTinterwordspacing
C.~J. Gaudet and A.~S. Maida, ``Deep quaternion networks,'' in
  \emph{IJCNN}.\hskip 1em plus 0.5em minus 0.4em\relax IEEE, 2018, pp. 1--8.
  [Online]. Available: \url{https://doi.org/10.1109/ijcnn.2018.8489651}
\BIBentrySTDinterwordspacing

\bibitem{pavllo2018quaternet}
D.~Pavllo, D.~Grangier, and M.~Auli, ``Quaternet: A quaternion-based recurrent
  model for human motion,'' in \emph{BMVC}, 2018.

\bibitem{pavllo2019modeling}
\BIBentryALTinterwordspacing
D.~Pavllo, C.~Feichtenhofer, M.~Auli, and D.~Grangier, ``Modeling human motion
  with quaternion-based neural networks,'' \emph{International Journal of
  Computer Vision}, pp. 1--18, 2019. [Online]. Available:
  \url{https://doi.org/10.1007/s11263-019-01245-6}
\BIBentrySTDinterwordspacing

\bibitem{ijcai2019-599}
\BIBentryALTinterwordspacing
S.~Zhang, L.~Yao, L.~Vinh~Tran, A.~Zhang, and Y.~Tay, ``Quaternion
  collaborative filtering for recommendation,'' in \emph{IJCAI}.\hskip 1em plus
  0.5em minus 0.4em\relax International Joint Conferences on Artificial
  Intelligence Organization, 7 2019, pp. 4313--4319. [Online]. Available:
  \url{https://doi.org/10.24963/ijcai.2019/599}
\BIBentrySTDinterwordspacing

\bibitem{tran2020quaternion}
\BIBentryALTinterwordspacing
T.~Tran, D.~You, and K.~Lee, ``Quaternion-based self-attentive long short-term
  user preference encoding for recommendation,'' in \emph{CIKM}, 2020, pp.
  1455--1464. [Online]. Available:
  \url{https://doi.org/10.1145/3340531.3411926}
\BIBentrySTDinterwordspacing

\bibitem{tay2019lightweight}
\BIBentryALTinterwordspacing
Y.~Tay, A.~Zhang, A.~T. Luu, J.~Rao, S.~Zhang, S.~Wang, J.~Fu, and S.~C. Hui,
  ``Lightweight and efficient neural natural language processing with
  quaternion networks,'' in \emph{ACL}.\hskip 1em plus 0.5em minus 0.4em\relax
  Florence, Italy: Association for Computational Linguistics, Jul. 2019, pp.
  1494--1503. [Online]. Available:
  \url{https://www.aclweb.org/anthology/P19-1145}
\BIBentrySTDinterwordspacing

\bibitem{parcollet2016quaternion}
\BIBentryALTinterwordspacing
T.~Parcollet, M.~Morchid, P.-M. Bousquet, R.~Dufour, G.~Linar{\`e}s, and
  R.~De~Mori, ``Quaternion neural networks for spoken language understanding,''
  in \emph{2016 IEEE Spoken Language Technology Workshop (SLT)}.\hskip 1em plus
  0.5em minus 0.4em\relax IEEE, 2016, pp. 362--368. [Online]. Available:
  \url{https://doi.org/10.1109/slt.2016.7846290}
\BIBentrySTDinterwordspacing

\bibitem{parcollet2019quaternion}
T.~Parcollet, M.~Ravanelli, M.~Morchid, G.~Linar{\`e}s, C.~Trabelsi,
  R.~De~Mori, and Y.~Bengio, ``Quaternion recurrent neural networks,'' in
  \emph{ICLR}, 2019.

\bibitem{bradbury2017quasi}
J.~Bradbury, S.~Merity, C.~Xiong, and R.~Socher, ``Quasi-recurrent neural
  networks,'' in \emph{ICLR}, 2017.

\bibitem{howard2018universal}
\BIBentryALTinterwordspacing
J.~Howard and S.~Ruder, ``Universal language model fine-tuning for text
  classification,'' in \emph{ACL}.\hskip 1em plus 0.5em minus 0.4em\relax
  Melbourne, Australia: Association for Computational Linguistics, Jul. 2018,
  pp. 328--339. [Online]. Available:
  \url{https://www.aclweb.org/anthology/P18-1031}
\BIBentrySTDinterwordspacing

\bibitem{mou2020malicious}
\BIBentryALTinterwordspacing
G.~Mou and K.~Lee, ``Malicious bot detection in online social networks: Arming
  handcrafted features with deep learning,'' in \emph{International Conference
  on Social Informatics}.\hskip 1em plus 0.5em minus 0.4em\relax Springer,
  2020, pp. 220--236. [Online]. Available:
  \url{https://doi.org/10.1007/978-3-030-60975-7_17}
\BIBentrySTDinterwordspacing

\bibitem{yao2018rdeepsense}
\BIBentryALTinterwordspacing
S.~Yao, Y.~Zhao, H.~Shao, A.~Zhang, C.~Zhang, S.~Li, and T.~Abdelzaher,
  ``Rdeepsense: Reliable deep mobile computing models with uncertainty
  estimations,'' \emph{ACM on Interactive, Mobile, Wearable and Ubiquitous
  Technologies}, vol.~1, no.~4, pp. 1--26, 2018. [Online]. Available:
  \url{https://doi.org/10.1145/3161181}
\BIBentrySTDinterwordspacing

\bibitem{hui2021evaluation}
\BIBentryALTinterwordspacing
L.~Hui and M.~Belkin, ``Evaluation of neural architectures trained with square
  loss vs cross-entropy in classification tasks,'' in \emph{ICLR}.\hskip 1em
  plus 0.5em minus 0.4em\relax OpenReview.net, 2021. [Online]. Available:
  \url{https://openreview.net/forum?id=hsFN92eQEla}
\BIBentrySTDinterwordspacing

\bibitem{kim2014convolutional}
\BIBentryALTinterwordspacing
Y.~Kim, ``Convolutional neural networks for sentence classification,'' in
  \emph{EMNLP}.\hskip 1em plus 0.5em minus 0.4em\relax Doha, Qatar: Association
  for Computational Linguistics, Oct. 2014, pp. 1746--1751. [Online].
  Available: \url{https://www.aclweb.org/anthology/D14-1181}
\BIBentrySTDinterwordspacing

\bibitem{indurthi2019fermi}
\BIBentryALTinterwordspacing
V.~Indurthi, B.~Syed, M.~Shrivastava, M.~Gupta, and V.~Varma, ``Fermi at
  {S}em{E}val-2019 task 6: Identifying and categorizing offensive language in
  social media using sentence embeddings,'' in \emph{International Workshop on
  Semantic Evaluation}.\hskip 1em plus 0.5em minus 0.4em\relax Minneapolis,
  Minnesota, USA: Association for Computational Linguistics, Jun. 2019, pp.
  611--616. [Online]. Available:
  \url{https://www.aclweb.org/anthology/S19-2109}
\BIBentrySTDinterwordspacing

\bibitem{devlin2019bert}
J.~Devlin, M.~Chang, K.~Lee, and K.~Toutanova, ``{BERT:} pre-training of deep
  bidirectional transformers for language understanding,'' in \emph{NAACL},
  2019.

\bibitem{morris2020textattack}
\BIBentryALTinterwordspacing
J.~Morris, E.~Lifland, J.~Y. Yoo, J.~Grigsby, D.~Jin, and Y.~Qi, ``Textattack:
  A framework for adversarial attacks, data augmentation, and adversarial
  training in nlp,'' in \emph{EMNLP}, 2020. [Online]. Available:
  \url{https://doi.org/10.18653/v1/2020.emnlp-demos.16}
\BIBentrySTDinterwordspacing

\bibitem{holtzman2020curious}
A.~Holtzman, J.~Buys, L.~Du, M.~Forbes, and Y.~Choi, ``The curious case of
  neural text degeneration,'' in \emph{ICLR}, 2020.

\bibitem{warstadt2019neural}
\BIBentryALTinterwordspacing
A.~Warstadt, A.~Singh, and S.~R. Bowman, ``Neural network acceptability
  judgments,'' \emph{Transactions of the Association for Computational
  Linguistics}, vol.~7, pp. 625--641, 2019. [Online]. Available:
  \url{https://doi.org/10.1162/tacl_a_00290}
\BIBentrySTDinterwordspacing

\bibitem{flesch1979write}
R.~Flesch, ``How to write plain english: Let’s start with the formula,''
  \emph{University of Canterbury}, 1979.

\bibitem{boughorbel2017optimal}
\BIBentryALTinterwordspacing
S.~Boughorbel, F.~Jarray, and M.~El-Anbari, ``Optimal classifier for imbalanced
  data using matthews correlation coefficient metric,'' \emph{PloS one},
  vol.~12, no.~6, 2017. [Online]. Available:
  \url{https://doi.org/10.1371/journal.pone.0177678}
\BIBentrySTDinterwordspacing

\end{thebibliography}

\end{document}